\documentclass[10pt,letterpaper]{article}
\usepackage[margin=1in]{geometry}
\usepackage{amsmath,amssymb}

\usepackage{changepage}

\usepackage{textcomp,marvosym}

\usepackage{cite}

\usepackage{nameref,hyperref}

\usepackage[right]{lineno}
\usepackage{setspace}

\usepackage[nopatch=eqnum]{microtype}
\DisableLigatures[f]{encoding = *, family = * }

\usepackage[table]{xcolor}

\usepackage{array}

\usepackage{float}

\newcolumntype{+}{!{\vrule width 2pt}}

\newlength\savedwidth

\usepackage[aboveskip=1pt,labelfont=bf,labelsep=period,justification=raggedright,singlelinecheck=off]{caption}

\makeatletter
\renewcommand{\@biblabel}[1]{\quad#1.}
\makeatother

\usepackage{lastpage,fancyhdr,graphicx}
\usepackage{epstopdf}
\fancyheadoffset[L]{2.25in}
\fancyfootoffset[L]{2.25in}

\usepackage{booktabs}
\usepackage{tabularx}
\usepackage{array}
\usepackage{placeins}

\begin{document}


\begin{center}
{\Large
\textbf{CEMIR vs. FISER: a comparative study of reasoning-based and enrichment-based approaches to pragmatic intent recognition}
}
\end{center}



\begin{center}

Ziya Hosseynzadeh\textsuperscript{1*},
Farjad Fallah\textsuperscript{2},
Farnoosh Fallah\textsuperscript{2},
Soroush Gooran\textsuperscript{3},
Mohammad Amanloo\textsuperscript{2},
Hossein Sameti\textsuperscript{3},
\par
\bigskip
\noindent
{\parindent=0pt}

\textbf{1} Speech and Language Processing Laboratory (SLPL), Sharif University of Technology, Tehran, Iran
\\
\textbf{2} School of Electrical and Computer Engineering, University of Tehran, Iran
\\
\textbf{3} Department of Computer Engineering, Sharif University of Technology, Tehran, Iran
\\
\bigskip
* ziyahosseynzadeh1@gmail.com
\end{center}

\section*{Abstract}
Intent recognition in human-robot interaction requires agents to resolve ambiguous natural language instructions by inferring the speaker's underlying goals. Research has increasingly focused on two notable paradigms for addressing this task: structured symbolic reasoning and multimodal input enrichment. However, it remains unclear whether enrichment-based methods provide benefit in settings where the ambiguity concerns unstated human intent rather than missing observable scene information. In this paper, we present a comparative and diagnostic study benchmarking FISER—a framework utilizing structured social and embodied reasoning—against CEMIR (Context-Enriched Multimodal Intent Recognition), a method that performs multimodal enrichment by synthesizing narratives and photorealistic images from textual prompts. By evaluating both approaches across four Large Language Models (LLMs), ranging from Qwen2.5‑Omni‑3B to GPT-5, on the HandMeThat v2 benchmark, we analyze their capability to resolve pragmatic ambiguities. Our statistical analysis, employing multiple-comparison correction, reveals no statistically significant difference in performance between FISER and CEMIR across any model or difficulty level. This pattern holds across all sixteen model-by-difficulty comparisons, as well as in a targeted ablation isolating CEMIR's narrative and visual components. Instead, performance is primarily associated with backbone capability and the inherent difficulty of the ambiguity. These findings suggest a boundary condition for text-derived multimodal enrichment: when the relevant task constraints are already available in the textual instruction, synthetic multimodal augmentation may not provide measurable benefit over structured reasoning on the same text.

\section*{Introduction}
Interactive AI systems such as smart home assistants, service robots, and virtual chatbots must infer user intent to act reliably and safely in everyday environments \cite{vanhorenbeke2021activity}. Intent recognition, the task of identifying the goal behind a user's instruction or action, is therefore a foundational component of natural language understanding in interactive settings, where misinterpretations can cause incorrect actions, breakdowns in collaboration, or degraded user trust \cite{zhou2024enhancing, chandrakala2024intent, benayas2024enhancing}. In domestic and assistive scenarios, users frequently omit details and rely on shared context, routines, and unspoken goals \cite{suchman1987plans}; consequently, brief utterances such as "Could you pass that?" can be under-specified without additional grounding \cite{clark1991grounding}. As communicative ambiguity increases, maintaining fluent and dependable human-AI/robot collaboration requires mechanisms that connect what is said to what is meant, and ultimately to what should be done \cite{lemaignan2012grounding}.  \newline  \newline
Meeting this requirement raises a basic design question that recurs across language understanding tasks more broadly: where should a system look to resolve what remains unclear? Prior work differs in its answer. One line of work treats the missing information as already recoverable from what is given to the agent, the human's prior actions and the language of the instruction itself, and resolves ambiguity through explicit, structured reasoning over that input; FISER \cite{Wan2025} exemplifies this approach on the HandMeThat benchmark \cite{wan2022handmethat}, decomposing intent recognition into a social reasoning stage and an embodied reasoning stage, and demonstrating strong performance using purely textual episode representations. A second line of work instead enriches the model's input with additional context before reasoning takes place, whether directly observed from the environment or, more recently, generated from the text itself, and has an established record of small but consistent gains on other language understanding tasks such as classification and QA \cite{tan2020vokenization, yang2022zlavi, lu2022imagination, huang2026synthetic}.  \newline  \newline
These two lines of work have developed largely independently and have not been evaluated under conditions the other is built for: reasoning-based approaches have been tested on tasks like HandMeThat \cite {wan2022handmethat} where ambiguity is pragmatic, the environment is fully described, and what remains uncertain is the human's intent, while enrichment-based approaches have been tested only on tasks where the missing information is a directly observable property of a static scene, or on single-step classification and QA judgments (Research Gap section). HandMeThat v2 offers a setting where this comparison becomes possible, and one where a strong reasoning-based baseline, FISER, already exists. The benchmark's textual interface deliberately withholds nothing perceptual: every object and its relevant properties are already stated in the episode text; this also means that any additional signal supplied to the model can only be self-generated, not genuinely external to the text, isolating the specific form of enrichment this study investigates from perceptual grounding based on direct observation.  \newline  \newline
The present study investigates whether an enrichment-based alternative can match or exceed FISER under this kind of ambiguity. We introduce CEMIR (Context-Enriched Multimodal Intent Recognition), which augments each action step of a HandMeThat v2 episode with a generated narrative and a corresponding generated image, produced entirely from the episode text rather than from any independent observation of the environment. The decision to combine narrative and visual elaboration jointly, rather than testing either in isolation, is motivated by Winston's Strong Story and Directed Perception Hypotheses, which hold that narrative plays a central role in human intelligence, and that much of human commonsense is derived by deploying perceptual apparatus on real and imagined events \cite{winston2011strong}; we adopt this as a design rationale for CEMIR's approach to resolving referential ambiguity, not as a claim on Winston's part about generated modalities or ambiguity resolution specifically. We evaluate CEMIR against FISER on 400 episodes from HandMeThat v2, across four language models ranging from 3B to GPT-5 scale, using a cluster-level paired permutation test with Holm–Bonferroni correction applied within each family of comparisons. Because CEMIR and FISER differ in both prompting structure and input modality, each is evaluated as a complete system representing its own paradigm; Ablation Study section separately reports how CEMIR's own performance changes when its narrative or image component is removed, to characterize their individual contributions within CEMIR's architecture. 
This work makes two contributions. First, we introduce CEMIR, an enrichment-based alternative to structured text-only reasoning for object and goal identification under pragmatic ambiguity. Second, we provide an empirical comparison between FISER and CEMIR across model scales and difficulty levels, together with a within-system ablation of CEMIR's narrative and visual components, addressing a comparison that, to our knowledge, existing work on synthetic multimodal augmentation has not examined.

\section*{Related Work}
\subsection*{Reasoning-Based Intent Recognition}
The inferential trajectory draws its foundations from Theory of Mind (ToM) research \cite{Byom2013ToM}, which models how agents infer the beliefs, desires, and intentions of others from observable behavior \cite{Baker2017ToM}. Bayesian Theory of Mind (BToM) \cite{Baker2011BToM} formalized this by treating human behavior as the outcome of rational planning, using Bayesian inference to recover latent intentions from observed actions. Hierarchical Bayesian Theory of Mind (HBToM) extended this framework by placing hierarchical priors over agents' goals and behavioral dispositions across multiple observed episodes, enabling few-shot inference of an agent's preferences from limited interaction history \cite{zhixuan2022solving}. These probabilistic formulations were frequently paired with Partially Observable Markov Decision Processes (POMDPs), enabling robots to reason about uncertain human intentions as evidence accumulates over time rather than from a single observation \cite{Bandyopadhyay2012}. A limitation shared by these Bayesian and POMDP-based approaches is their dependence on hand-specified goal spaces and transition models, which restricts their applicability to open-ended natural language instructions \cite{tan2023language}. \newline

Addressing this limitation, more recent work has incorporated language directly into the inverse planning process, and increasingly relies on large language models rather than hand-specified goal spaces. CLIPS (Cooperative Language-Guided Inverse Plan Search) jointly conditions goal inference on both observed actions and ambiguous instructions, using an LLM to score the likelihood of an instruction under each hypothesized plan within a sequential inverse planning framework \cite{zhixuan2024clips}. Tomcat instead departs from this Bayesian formulation entirely, using few-shot chain-of-thought prompting to interpret incomplete or ambiguous instructions and infer a human principal's underlying intent; on a purpose-built instruction inference task, it matches the performance of human participants \cite{saad2025theory}. A related approach disambiguates underspecified language commands by prompting an LLM to reconcile a human demonstration with a reference trajectory, recovering the goal that the instruction alone leaves unstated \cite{hwang2025maskedirl}. FISER (Follow Instructions with Social and Embodied Reasoning) exemplifies this same shift toward language-native reasoning pipelines built directly on LLMs, decomposing intent recognition into two stages: a social reasoning stage that infers the human's underlying goal from contextual cues and task semantics, and an embodied reasoning stage that translates the inferred goal into a feasible action sequence \cite{Wan2025}. FISER is positioned against end-to-end instruction-to-action mapping and Chain-of-Thought prompting baselines, and its evaluations rely exclusively on textual episode representations of the HandMeThat benchmark \cite{Wan2025, wan2022handmethat}.Across this line of work, from Bayesian goal inference to LLM-based reasoning pipelines, ambiguity is resolved entirely through inference over what is already given to the agent: prior actions and instruction text. None of these approaches incorporates a perceptual signal, generated or observed, external to this input.

\subsection*{Enrichment-Based Approaches to Ambiguity Resolution}
A second line of work addresses instruction ambiguity not by making the inferential process more explicit, but by enriching the model's input with additional context before reasoning takes place. This strategy has taken two distinct forms, depending on whether the added content comes from genuine, independent access to the environment, or is instead synthesized or retrieved without any such access.

\subsubsection*{Grounding in Directly Observed Perceptual Signals}

Language alone frequently underdetermines which object, location, or action a human intends, and robots operating in real environments have direct access to visual and contextual signals that can resolve this underdetermination \cite{tellex2011understanding}. An early formalization of this idea is the Multi-Modal Classifier Generative Adversarial Network (MMC-GAN), which addresses carry-and-place instructions in which the target area is left unspecified (e.g., "put away the milk and cereal") by conditioning target prediction jointly on the instruction, the robot's physical state, and the clutter of candidate destinations in the environment \cite{magassouba2018multimodal}. Subsequent work has extended the range of such signals to embodied cues produced by the human during the interaction itself: SemanticScanpath combines a user's gaze behavior with speech to disambiguate spoken requests, using an LLM-based agent to jointly interpret where the human is looking and what they are saying \cite{ menendez2025semanticscanpath}. PaLM-E integrates such grounding directly into a single end-to-end model, connecting a large language model with real-world sensor modalities, images, robot state, and other continuous observations, encoded as tokens interleaved with text \cite{driess2023palme}. Across this progression, the disambiguating signal always originates from direct observation of the physical world, a camera feed, a gaze tracker, a robot's proprioceptive state, rather than from the instruction text itself; the perceptual channel supplies information the text does not already contain.

\subsubsection*{Enrichment from Self-Generated or Retrieved Content}

A separate strand of work asks a narrower question: can a signal that carries no independent access to the environment, only content derived from or retrieved alongside the text itself, still improve a model's handling of that text? Tan and Bansal pre-train a language model using images retrieved for each word from an independent image-caption corpus as auxiliary supervision, and show this improves performance on text-only benchmarks such as GLUE and SQuAD, even though no images are used at test time \cite{tan2020vokenization}. More directly, images synthesized entirely from the input text, with no independent access to the environment at all, have also been shown to help: Yang et al. report consistent zero-shot gains from generated visual imagination across several classification and science-QA benchmarks \cite{yang2022zlavi}, and Lu et al. report similar gains from a generation-based imagination encoder in few-shot natural language understanding on SWAG \cite{lu2022imagination}. Huang et al. similarly fuse a generated image with its source text for text-centric classification, finding small but consistent accuracy gains that track how visually descriptive the source text is to begin with \cite{huang2026synthetic}.  \newline

Unlike the perceptual grounding methods above, none of these approaches has access to any signal beyond what the source text already encodes; any benefit they produce cannot be attributed to genuinely new information. Whether such enrichment is theoretically well-motivated is, on this basis, contested: the classical grounding hypothesis holds that a symbol cannot acquire meaning through reference to other symbols in the same representational system alone \cite{harnad1990symbol}, and prominent recent work argues that language understanding is fundamentally limited without a perceptual or embodied signal external to the text \cite{bisk2020experience}. Under this view, content generated entirely from a piece of text should carry no information beyond what the text already encodes, yet the empirical results above show a measurable, if modest, benefit, at least in single-step classification and QA tasks.

\subsection*{Research Gap}

The two lines of work reviewed above have developed largely independently of one another, and neither has been evaluated under the conditions the other is built for. Reasoning-based approaches, including CLIPS, Tomcat, Masked IRL, and FISER, resolve ambiguity entirely through inference over text or symbolic state, without any perceptual signal, and have been evaluated in settings, HandMeThat among them, where the difficulty is pragmatic: the environment is fully described, and what remains uncertain is the human's intent \cite{Wan2025, wan2022handmethat, zhixuan2024clips, saad2025theory, hwang2025maskedirl}. Enrichment-based approaches, whether grounded in directly observed perceptual signals \cite{magassouba2018multimodal, menendez2025semanticscanpath, driess2023palme} or in content generated or retrieved without independent access to the environment \cite{tan2020vokenization, yang2022zlavi, lu2022imagination, huang2026synthetic}, have instead been evaluated on tasks where the missing information is a directly observable property of a static scene, or on single-step classification and QA judgments, never on a task where the missing information is a person's unstated intent.  \newline

HandMeThat v2 is, to our knowledge, one of the first settings in which this comparison becomes possible: its textual interface is deliberately constructed to exclude visual perception and object recognition, so that every object and its relevant properties are already stated in the episode text, and what remains difficult is inferring, from a human's prior actions and an underspecified instruction, which object or goal is meant \cite{wan2022handmethat}. A strong reasoning-based baseline, FISER, already exists for this benchmark \cite{Wan2025}. What has not been examined is whether an enrichment-based system, one that supplies the model with self-generated context rather than relying on explicit structured reasoning, can match or exceed this reasoning-based approach under pragmatic ambiguity. We address this gap directly by introducing CEMIR and comparing it against FISER under identical experimental conditions, across four model scales and four ambiguity levels, with a within-system ablation isolating the contribution of CEMIR's narrative and visual components (Ablation Study section).

\section*{Proposed Method}

\subsection*{CEMIR Framework}
CEMIR (Context-Enriched Multimodal Intent Recognition) serves as the multimodal configuration evaluated in this study. The central idea is to enrich the same textual episode used in the reasoning-based baseline with additional perceptual and narrative context to reduce ambiguity in intent interpretation. As noted, the rationale for pairing narrative and visual enrichment, rather than testing either alone, comes from Winston's Strong Story and Directed Perception Hypotheses \cite{winston2011strong}: the view that narrative is central to human intelligence, and that people derive much of their commonsense by perceptually engaging with both real and imagined events. We treat this only as inspiration for CEMIR's architecture, not as an empirical claim by Winston about generated content or about ambiguity resolution specifically. To produce these modalities, we implemented an enrichment pipeline using Qwen3-8B \cite{qwen3} as the text generator and FLUX.1 [dev] as the image generator. Each action description is processed independently at the step level rather than the episode level, allowing the model to preserve localized contextual cues such as object attributes, spatial relations, and interaction affordances. As a result, for an episode containing N actions, the enrichment process produces N narrative descriptions and N corresponding visual frames (one multimodal pair per action step). \newline  \newline
 For each action step, a generator model (Qwen3-8B) produces a literal scene description  from the environment and the action history. This description serves two roles: it is provided to the backbone model as text (the narrative modality), and it is used as the prompt for an image generator (FLUX.1 [dev] \cite{Flux2024}) to synthesize a 512×512 image (the visual modality). Negative prompts are applied during generation to suppress unrealistic or visually irrelevant artifacts. The complete enrichment dataset (generated narratives and images) is publicly available. All enriched outputs are stored in a machine-readable JSON format aligned with the original dataset schema to ensure full reproducibility. During inference, each episode is therefore represented as a multimodal sequence consisting of the original textual trajectory together with the generated narrative and visual frame for every action step. These modalities are provided jointly to the evaluated models through their native vision–language interfaces. By augmenting the original textual trajectory with structured perceptual and narrative context, CEMIR enables a direct comparison between multimodal input enrichment and the reasoning-based FISER prompting protocol under otherwise identical experimental conditions. The overall architecture of the CEMIR framework is illustrated in figure 1.

\begin{figure}[H]
\centering
\includegraphics[width=\textwidth]{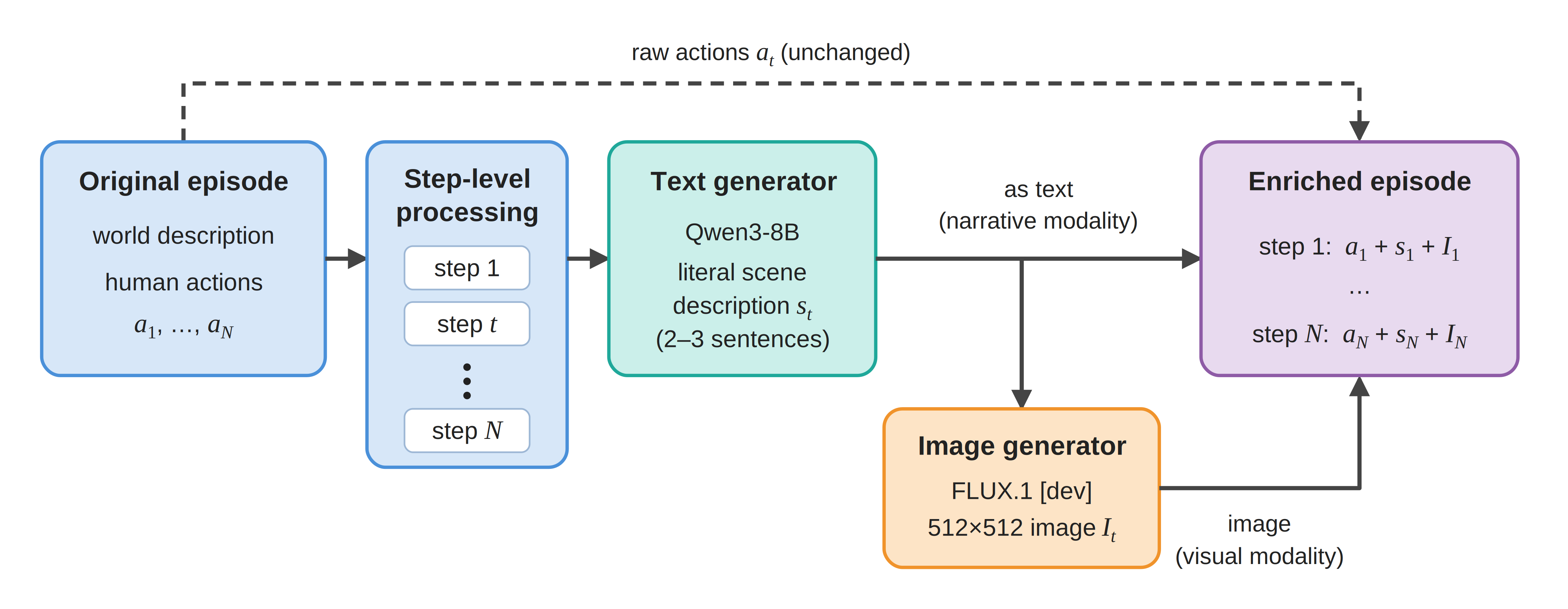}

\caption{The CEMIR enrichment pipeline. For each action step $t$, a scene description $s_t$ (Qwen3-8B) is used both as text and as the prompt for an image $I_t$ (FLUX.1 [dev]); together with the unchanged raw action $a_t$, they form the enriched step.}
\label{fig:cemir-framework}
\end{figure}

\section*{Dataset and Experimental Setup}
\subsection*{Dataset}
We evaluate both frameworks on HandMeThat v2 \cite{wan2022handmethat}, a benchmark for ambiguous instruction following in text-based human–robot interaction environments. Each episode describes a household scenario in which a human agent performs a sequence of actions toward an implicit goal and then issues an underspecified natural language instruction to a robot agent. The robot must infer the human's intent within a limited interaction budget. The dataset organizes episodes into four ambiguity levels of increasing difficulty: Level 1 requires literal grounding with no ambiguity; Level 2 requires inferring the human's goal from prior actions; Level 3 additionally requires pragmatic reasoning about language use; and Level 4 involves inherent ambiguity that cannot be fully resolved without strong priors over human preferences \cite{wan2022handmethat}. To ensure balanced evaluation while maintaining computational tractability, we constructed a structured subset of the dataset comprising 400 episodes in total. Specifically, we selected 4 episodes per (goal template, difficulty level) cell, across 25 goal templates and 4 ambiguity levels (25 × 4 × 4 = 400). This sampling strategy ensures an equal distribution of 100 episodes per ambiguity level, providing a rigorous and balanced comparison across the different difficulty conditions.
\subsection*{Baseline (FISER Framework)}
FISER (Follow Instructions with Social and Embodied Reasoning) \cite{Wan2025}, described in Reasoning-Based Intent Recognition Section, serves as the reasoning-based baseline in this study. The original work implements both a Transformer-based model trained from scratch and a prompted LLM variant using GPT-4 Turbo with Chain-of-Thought reasoning \cite{Wan2025}. In the present study, we adopt the prompting protocol of the LLM-based variant while replacing the backbone with the evaluated models (GPT-5, GPT-5-Mini, GPT-5-Nano, and Qwen2.5-Omni-3B). This choice aligns with our research objective of comparing reasoning-based and multimodal configurations across different model capacities rather than training specialized architectures. FISER inputs consist of the raw textual episode (world state description, human trajectory, and natural language instruction), formatted according to the original prompting protocol \cite{Wan2025}. Exemplar selection is described in Models and Configurations Section.

\subsection*{Content Fidelity Verification}
Two independent annotators reviewed a stratified sample of 40 episodes (10 per difficulty level), covering 498 action steps, rating the generated narrative and image for each step as Pass or Fail against the episode's structured ground-truth data (e.g., object states and the action sequence); task-relevant details had to be consistent with the ground truth for a Pass, while stylistic or non-task-relevant embellishments (e.g., unspecified spatial arrangements, background elements, or visual style) were not counted against a rating. The two annotators agreed on 96\% of ratings; the small number of disagreements were resolved through discussion. Under this protocol, 469 of 498 audited action steps (94.18\%) were rated Pass, indicating that the generated content was, in the large majority of cases, consistent with the underlying episode state. A small number of exceptions were noted, most commonly involving a location or state misattributed to an earlier or later point in the action sequence.  

\subsection*{Models and Configurations}

Both frameworks were evaluated across four language models representing different reasoning capacities and computational scales: Qwen2.5‑Omni‑3B, GPT‑5‑Nano, GPT‑5‑Mini, and GPT‑5. Qwen2.5‑Omni‑3B is an open‑weight language model representing small‑scale reasoning under minimal computational cost \cite{qwen25omni}. GPT‑5‑Nano and GPT‑5‑Mini are compact commercial models providing moderate reasoning capacity, while GPT‑5 represents the upper bound in reasoning and representational capability among the evaluated systems. This selection covers a broad spectrum of model scales, ranging from lightweight open models to state‑of‑the‑art commercial systems while remaining computationally feasible. All models were evaluated under a two-shot inference protocol. For each evaluation episode, two solved exemplars were sampled at random from the 400-episode subset, excluding the current instance; sampling was performed independently for each condition and run. In the CEMIR condition, exemplars follow the same multimodal input format used during inference, whereas FISER exemplars are provided as textual trajectories.

\subsection*{Evaluation Protocol and Metrics}

Each evaluation run followed a single-attempt protocol: for each episode, the model was given one opportunity to produce an action sequence within a fixed horizon of seven steps, without retries, rollouts, or corrective feedback. This protocol is consistent with the evaluation setting adopted in the original FISER paper \cite{Wan2025}, ensuring direct comparability between the two frameworks. All language models were evaluated using the default decoding configurations provided by their respective APIs. To account for stochasticity in model outputs, each episode was evaluated across three independent runs for each model. Seven of the 7,200 attempts across the GPT-5-family conditions ($<0.1\%$) returned no valid output and were excluded; these fell in five model × difficulty level × method cells (listed in the note to Table 3), and success rates for the affected cells are computed over the remaining attempts. Success rate is defined as the percentage of episodes in which the model correctly identifies the ground-truth target object within the step budget. Results are reported separately for each model across four ambiguity levels. \newline

To assess whether observed differences between FISER and CEMIR are statistically significant, we treated each episode as the independent unit of analysis: for each episode, the per-method outcome was computed as the mean over its three repetitions, and paired differences between methods were tested via a cluster-level, sign-flip permutation test (episode as the independent unit, 20{,}000 permutations), with 95\% confidence intervals obtained via a nonparametric bootstrap (2{,}000 resamples, percentile method; resampling episodes), and Holm--Bonferroni correction (\texttt{statsmodels} v0.14.2, \texttt{multipletests}, method="holm") applied independently within each table (Tables 1--4); Table 3, for instance, comprises 16 model-level comparisons (four backbones × four difficulty levels). The significance threshold was set at $\alpha = 0.05$. Results are reported as the difference in accuracy between CEMIR and FISER in percentage points, computed as the mean of per-episode differences, along with cluster-bootstrap 95\% confidence intervals and Holm-corrected p-values. All analyses were performed in Python 3.12.7 using pandas (v2.2.2) and NumPy (v1.26.4); a fixed random seed (seed = 42) was used throughout for reproducibility. All evaluation and analysis code is publicly available.

\subsection*{Modality Ablation}

To isolate the individual contribution of each generated modality within CEMIR, we additionally evaluated two reduced variants of the framework on Qwen2.5‑Omni‑3B, selected due to its lower inference cost, which made the additional evaluation conditions computationally feasible, (Ablation Study section). In the Raw Text + Generated Narrative condition, each action step retains CEMIR's generated narrative paraphrase but omits the corresponding visual frame; in the Raw Text + Image condition, each action step retains the generated visual frame but omits the narrative paraphrase. Both variants use the same enrichment pipeline, prompting structure, and inference protocol as full CEMIR. Neither variant reintroduces FISER's social/embodied reasoning decomposition; both remain CEMIR-based systems evaluated against the FISER baseline under the same protocol described in Evaluation Protocol and Metrics section.
\section*{Results}
This section evaluates the impact of the proposed CEMIR framework on task success rates across multiple large language models and varying levels of environmental complexity. In particular, we examine whether enriching the agent’s contextual representation with multimodal information leads to performance improvements, or conversely introduces any degradation compared to the baseline FISER method. To this end, we conduct a paired evaluation across several LLM backbones and difficulty levels, measuring the Exact Success Rate (SR\%) at the episode level. Statistical comparisons between CEMIR and FISER are performed using a cluster-level paired permutation test, with Holm–Bonferroni correction applied to control for multiple comparisons.

\subsection*{Comparative Analysis Across LLM Backbones}
Table 1 and Fig. 2 present the aggregate success rates across all tested LLM backbones. Overall, CEMIR maintained a performance level highly consistent with the FISER baseline, with no statistically significant differences observed after the Holm-Bonferroni correction ($p_{\mathrm{holm}}>0.05$ for all models). For the most capable model, GPT-5, CEMIR achieved a success rate of 29.08\%, a marginal improvement over FISER's 28.67\% ($\Delta = 0.42\,\text{pp}$). A similar trend was observed for the lightweight Qwen2.5‑Omni‑3B model, where CEMIR outperformed FISER by $0.75\,\text{pp}$ (reaching 9.58\%). While these gains were not statistically significant, they indicate that the additional multimodal context did not measurably degrade performance at any model scale. Furthermore, the 95\% Confidence Intervals (CI) for the performance delta consistently bound zero for all models, reinforcing the conclusion that CEMIR shows no statistically significant difference from FISER across varying backbone capabilities, from small-scale parameters (3B) to state-of-the-art architectures.

\begin{table}[!ht]
\centering
\small
\setlength{\tabcolsep}{3pt}
\caption{Aggregate Success Rate (SR\%) comparison between FISER and CEMIR across different LLM backbones. $\Delta$ denotes CEMIR $-$ FISER. P-values from a cluster-level paired permutation test (episode as the independent unit) with Holm-Bonferroni correction; CIs are cluster-bootstrap 95\% intervals.}
\label{table1}
\begin{tabular}{
  >{\raggedright\arraybackslash}p{2.8cm}
  >{\centering\arraybackslash}p{0.5cm}
  >{\centering\arraybackslash}p{1.3cm}
  >{\centering\arraybackslash}p{1.3cm}
  >{\centering\arraybackslash}p{2.6cm}
  >{\centering\arraybackslash}p{1.6cm}
  >{\centering\arraybackslash}p{1.1cm}
  >{\centering\arraybackslash}p{1.3cm}
  >{\raggedright\arraybackslash}p{2.1cm}
}
\toprule
\textbf{Model} & \textbf{n} & \textbf{FISER SR\%} & \textbf{CEMIR SR\%} & \textbf{$\Delta$ exact\_acc (CEMIR$-$FISER)} & \textbf{95\% CI} & \textbf{$p$-value} & \textbf{$p$-value (Holm)} & \textbf{Significance} \\
\midrule
GPT-5 & 400 & 28.67 & 29.08 & 0.42 & [-1.25, 2.25] & 0.7147 & 1.0000 & N.S. (Not Significant) \\
GPT-5-MINI & 400 & 27.21 & 26.25 & -1.08 & [-2.83, 0.67] & 0.2489 & 0.9954 & N.S. \\
GPT-5-NANO & 400 & 19.83 & 19.27 & -0.58 & [-2.50, 1.33] & 0.5971 & 1.0000 & N.S. \\
Qwen2.5‑Omni‑3B & 400 & 8.83 & 9.58 & 0.75 & [-1.59, 3.00] & 0.5762 & 1.0000 & N.S. \\
\bottomrule
\end{tabular}

\vspace{5pt}

\begin{minipage}{\linewidth}
\small
\textbf{Note:} $\Delta$ is the mean of episode-level paired differences, whereas \textit{SR} columns are pooled over attempts; the two differ slightly for GPT-5-mini and GPT-5-nano, where a few attempts were missing.
\end{minipage}

\end{table}

\begin{figure}[!ht]
\centering
\includegraphics[width=\textwidth]{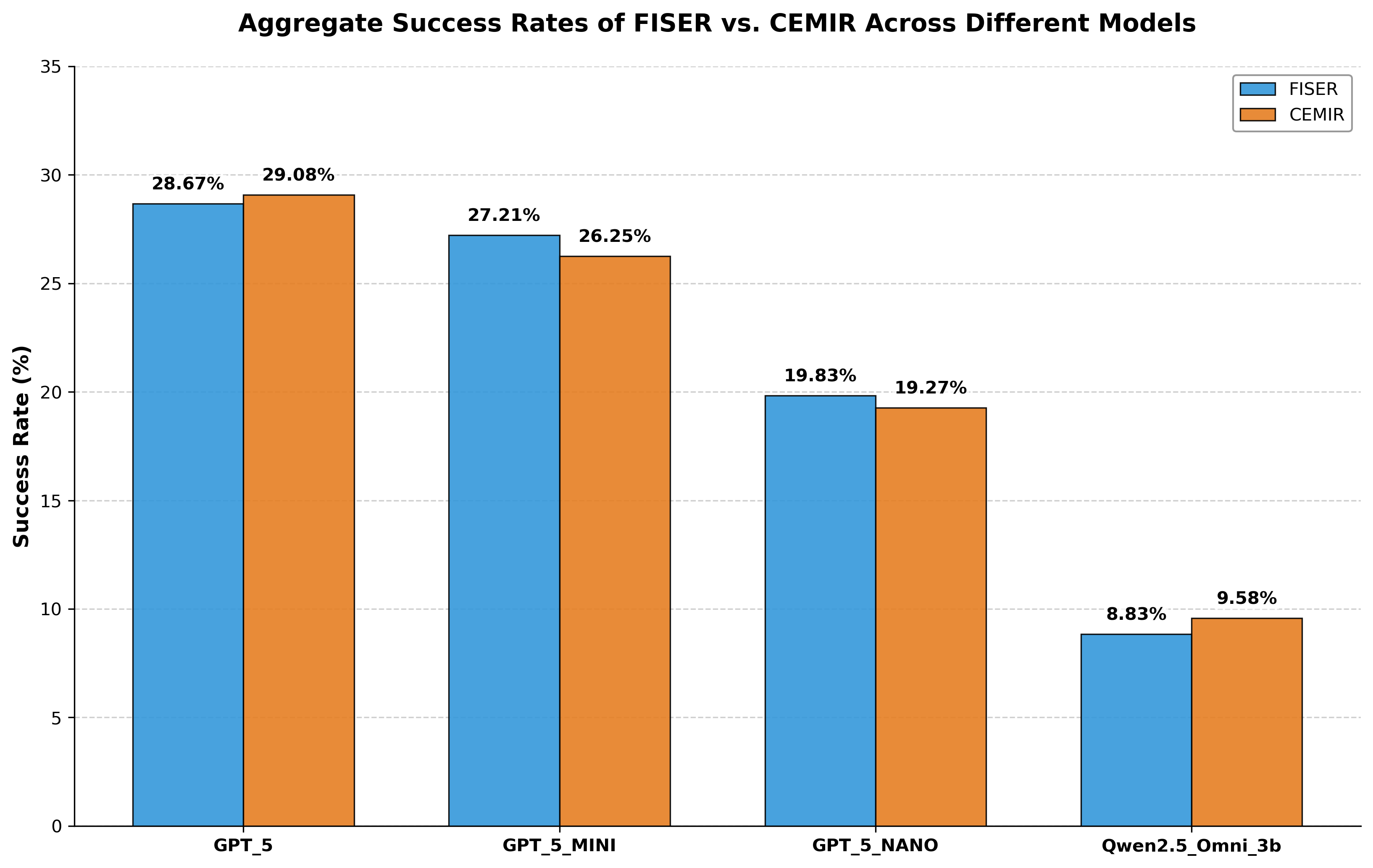}
\caption{Aggregate success rates of FISER vs. CEMIR across different models.}
\label{fig:FISER vs. CEMIR }
\end{figure}

\subsection*{Impact of Environmental Complexity}
We further examined the performance of both methods across four pre-defined difficulty levels. As shown in Table 2 and Figure 3, there is a clear inverse correlation between the environment's difficulty level and the Success Rate (SR\%) of both FISER and CEMIR, with SR dropping from approximately 60\% at Level 1 to below 6\% at Level 4. Statistical analysis (a cluster-level paired permutation test) was conducted to compare the two methods at each level:

\begin{itemize}
    \item \textbf{Level 1:} FISER achieved a slightly higher SR (60.88\%) compared to CEMIR (59.58\%), with a difference of $-1.31\,\text{pp}$ ($p_{\text{holm}}=1.00$).
    \item \textbf{Level 2:} CEMIR outperformed FISER by $1.58\,\text{pp}$ (13.17\% vs 11.58\%), although this difference was not statistically significant ($p_{\text{holm}}=1.00$).
    \item \textbf{Level 3:} A minimal difference of $-0.75\,\text{pp}$ was observed in favor of FISER ($p_{\text{holm}}=1.00$).
    \item \textbf{Level 4:} Both methods reached nearly identical Success Rates (FISER: 5.92\%, CEMIR: 5.94\%; $\Delta=0.02\text{ pp}$, $p_{\text{holm}}=1.00$).
\end{itemize}
These results indicate that the integration of CEMIR does not lead to a statistically significant change in performance, either improvement or degradation, across any of the tested difficulty tiers. This consistency, particularly the identical results in Level 4, suggests that the multimodal enrichment maintains the system's baseline reasoning stability across varying levels of environmental complexity.

\begin{table}[!ht]
\centering
\small
\setlength{\tabcolsep}{3pt}
\caption{Comparative Success Rates (SR\%) by difficulty level. $\Delta$ represents the performance margin (CEMIR $-$ FISER). All $p_{\mathrm{holm}}$ values exceed the 0.05 threshold.}
\label{table2}
\begin{tabular}{
  >{\centering\arraybackslash}p{1.2cm}
  >{\centering\arraybackslash}p{0.9cm}
  >{\centering\arraybackslash}p{1.3cm}
  >{\centering\arraybackslash}p{1.3cm}
  >{\centering\arraybackslash}p{2.6cm}
  >{\centering\arraybackslash}p{1.6cm}
  >{\centering\arraybackslash}p{1.1cm}
  >{\centering\arraybackslash}p{1.3cm}
  >{\raggedright\arraybackslash}p{2.1cm}
}
\toprule
\textbf{Level} & \textbf{n} & \textbf{FISER SR\%} & \textbf{CEMIR SR\%} & \textbf{$\Delta$ exact\_acc (CEMIR$-$FISER)} & \textbf{95\% CI} & \textbf{$p$-value} & \textbf{$p$-value (Holm)} & \textbf{Significance} \\
\midrule
1 & 100 & 60.88 & 59.58 & -1.31 & [-3.50, 0.80] & 0.2589 & 1.0000 & N.S. (Not Significant) \\
2 & 100 & 11.58 & 13.17 &  1.58 & [-1.08, 4.17] & 0.2766 & 1.0000 & N.S. \\
3 & 100 & 6.17  &  5.42 & -0.75 & [-1.92, 0.50] & 0.2822 & 1.0000 & N.S. \\
4 & 100 & 5.92  &  5.94 &  0.02 & [-1.50, 1.61] & 1.0000 & 1.0000 & N.S. \\
\bottomrule
\end{tabular}

\vspace{5pt}

\begin{minipage}{\linewidth}
\small
\textbf{Note:} Each level contains $K = 100$ independent test episodes evaluated across 4 LLM backbones ($n = 400$ total evaluations per level). Statistical significance was assessed at the episode cluster level ($K = 100$).
\end{minipage}
\end{table}

\begin{figure}[!ht]
\centering
\includegraphics[width=\textwidth]{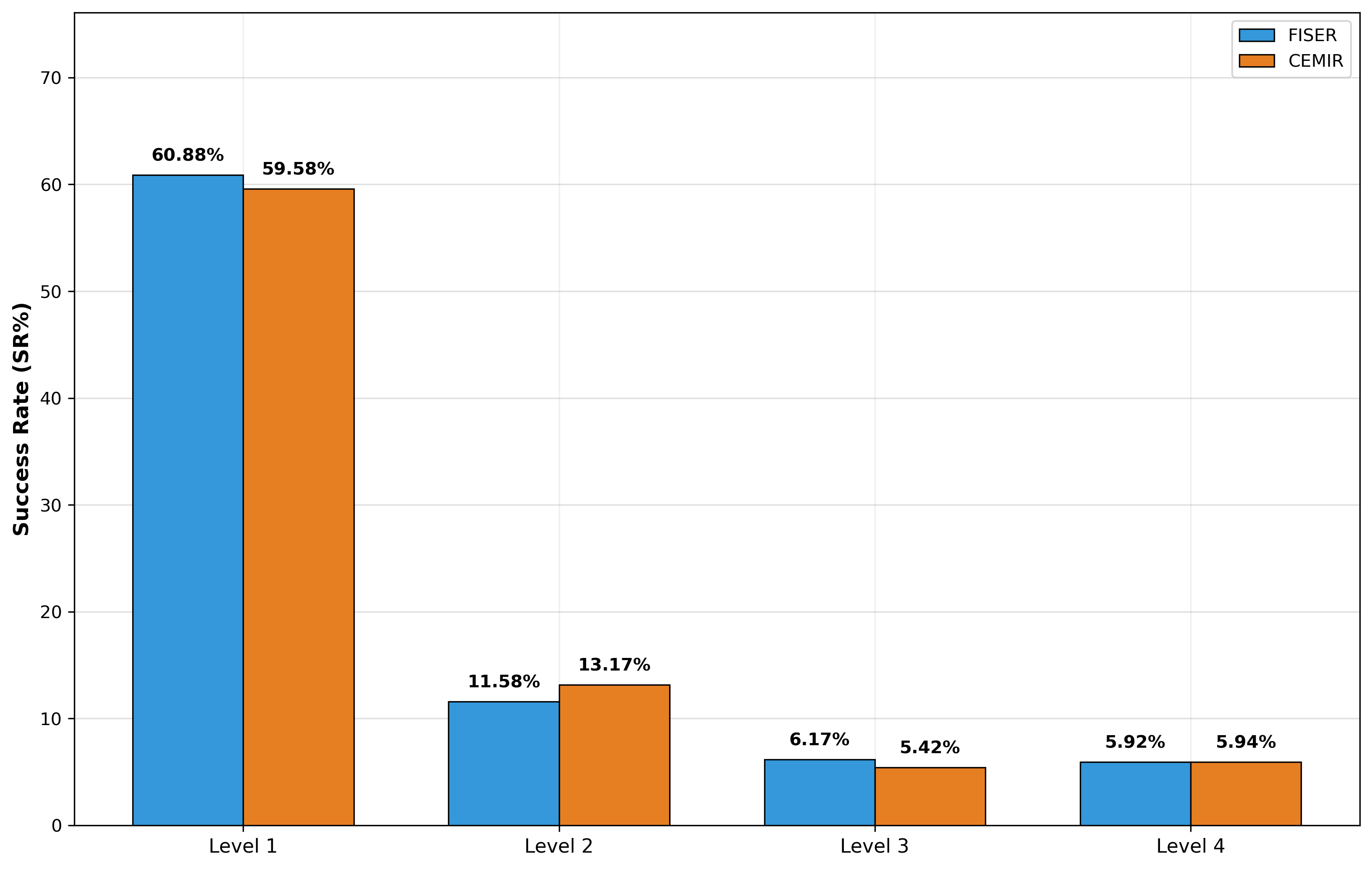}
\caption{Success Rate (SR\%) of FISER and CEMIR across four difficulty levels.}
\label{fig:FISER and CEMIR}
\end{figure}

\subsection*{Interaction Between Model Scale and Environmental Difficulty}

This section examines how the success rates of both FISER and CEMIR are influenced by the choice of LLM backbone and the level of environmental complexity. Table 3 and Figures 4–5 illustrate the performance of four models across all difficulty tiers, providing a comprehensive view of how these factors interact.

\subsubsection*{1. Performance Correlation with Backbone Capability:}
A consistent performance hierarchy is observed across all tested scenarios. For both FISER and CEMIR, transitioning from a lower-capability model (Qwen2.5‑Omni‑3B) to a higher-capability model (GPT-5) is associated with a substantial increase in success rates, particularly in low-complexity environments. For instance, at Level 1, the success rate for both methods increases from approximately 20\% to over 80\% as backbone capability increases. This highlights a strong positive correlation between backbone capability and the overall success of the agent in these tasks.

\subsubsection*{2. Sensitivity to Task Complexity Across Models:}

The data indicates that environmental difficulty impacts all models in a similar manner. Even for the most capable models like GPT-5, there is a marked decline in performance as complexity increases, with success rates dropping from approximately 83\% at Level 1 to approximately 8\% at Level 4. This downward trend is mirrored in both FISER and CEMIR across all model sizes, suggesting that the challenges posed by the environment affect the agents’ performance regardless of the underlying LLM or the specific method used.

\subsubsection*{3. Comparative Stability}
Across 16 distinct experimental conditions (4 models × 4 difficulty levels), the performance of CEMIR remains statistically indistinguishable from FISER. The fluctuations in success rates ($\Delta$) are minimal and do not follow a pattern of consistent improvement or degradation. This consistency suggests that the integration of multimodal features in CEMIR does not significantly alter the performance profile of the baseline across different model scales or complexity levels.

\subsection*{Ablation Study: Modality Contributions (Qwen2.5‑Omni‑3B)}

To examine the individual contribution of each modality within the CEMIR framework, we conducted a targeted ablation study using Qwen2.5‑Omni‑3B, the only model available under our computational constraints. Two modality conditions were evaluated against the FISER baseline: (1) Raw Text + Generated Narrative, which augments the original textual episode with step-level narrative paraphrases only, and (2) Raw Text + Image, which augments the episode with generated visual frames only. Each episode was run across three repetitions; because repetitions of the same episode are not independent observations, statistical comparisons treated the episode (n = 100 per cell) as the independent unit, using a cluster-level paired permutation test with cluster-bootstrap 95\% confidence intervals and Holm–Bonferroni correction, consistent with the primary analysis (Evaluation Protocol and Metrics section). Results are reported in Table 4. As shown in Table 4, neither modality condition produced a statistically significant improvement over FISER at any difficulty level. The one exception to this pattern at the unadjusted level is Raw Text + Image at Level 2 (+7.33~pp, unadjusted $p = 0.0090$, 95\% CI [2.00, 12.67]), which does not survive Holm--Bonferroni correction ($p_{\text{Holm}} = 0.0716$) and does not recur at the other three levels.

\section*{Analysis and Discussion}
The experimental results reveal a consistent pattern across all evaluated conditions: the performance of CEMIR remains statistically indistinguishable from that of the FISER baseline, while substantial variation in success rates is primarily associated with the choice of LLM backbone and the level of environmental difficulty. Across sixteen experimental configurations (four models × four difficulty levels), both methods exhibit nearly identical performance trajectories. Improvements in backbone capability are correlated with higher success rates, whereas increasing environmental complexity leads to systematic performance degradation for both approaches. This consistent alignment suggests that, within the evaluated benchmark, the dominant factors shaping task success are the reasoning capacity of the underlying language model and the intrinsic difficulty of the environment, rather than the choice between the two paradigms compared here. \newline

\begin{table}[!ht]
\centering
\small
\setlength{\tabcolsep}{3pt}
\caption {Detailed Success Rates (SR\%) and statistical significance for each LLM backbone across four difficulty levels.}
\label{table3}
\begin{tabular}{
  >{\raggedright\arraybackslash}p{2.4cm}
  >{\centering\arraybackslash}p{0.7cm}
  >{\centering\arraybackslash}p{0.8cm}
  >{\centering\arraybackslash}p{1.2cm}
  >{\centering\arraybackslash}p{1.2cm}
  >{\centering\arraybackslash}p{1.0cm}
  >{\centering\arraybackslash}p{1.6cm}
  >{\centering\arraybackslash}p{1.1cm}
  >{\centering\arraybackslash}p{1.1cm}
  >{\raggedright\arraybackslash}p{2.1cm}
}
\toprule
\textbf{Model} & \textbf{Level} & \textbf{n} & \textbf{FISER (\%)} & \textbf{CEMIR (\%)} & \textbf{$\Delta$ (pp)} & \textbf{95\% CI} & \textbf{$p$-value} & \textbf{$p$-value (Holm)} & \textbf{Significance} \\
\midrule
GPT-5 & 1 & 100 & 81.33 & 83.00 & 1.67 & [-1.67, 5.33] & 0.4936 & 1.0000 & N.S. (Not Significant) \\
      & 2 & 100 & 18.00 & 18.33 & 0.33 & [-4.67, 5.33] & 1.0000 & 1.0000 & N.S. \\
      & 3 & 100 & 8.00  & 7.00  & -1.00 & [-3.33, 1.00] & 0.5627 & 1.0000 & N.S. \\
      & 4 & 100 & 7.33  & 8.00  & 0.67 & [-1.67, 3.33] & 0.8078 & 1.0000 & N.S. \\
\cmidrule(lr){1-10}
GPT-5-MINI & 1 & 100 & 77.26 & 77.00 & -0.33 & [-3.67, 3.00] & 1.0000 & 1.0000 & N.S. \\
           & 2 & 100 & 15.00 & 16.00 & 1.00 & [-3.67, 5.67] & 0.7827 & 1.0000 & N.S. \\
           & 3 & 100 & 8.36  & 5.35  & -3.00 & [-5.67, -0.67] & 0.0441 & 0.7056 & N.S. \\
           & 4 & 100 & 8.33  & 6.40  & -2.00 & [-4.33, 0.00] & 0.1513 & 1.0000 & N.S. \\
\cmidrule(lr){1-10}
GPT-5-NANO & 1 & 100 & 63.33 & 58.33 & -5.00 & [-10.33, 0.00] & 0.0752 & 1.0000 & N.S. \\
           & 2 & 100 & 7.33  & 8.00  & 0.67 & [-2.67, 4.33] & 0.8592 & 1.0000 & N.S. \\
           & 3 & 100 & 4.33  & 5.67  & 1.33 & [-1.00, 3.67] & 0.4084 & 1.0000 & N.S. \\
           & 4 & 100 & 4.33  & 5.02  & 0.67 & [-2.33, 3.67] & 0.8271 & 1.0000 & N.S. \\
\cmidrule(lr){1-10}
Qwen2.5‑Omni‑3B & 1 & 100 & 21.67 & 20.00 & -1.67 & [-8.00, 4.67] & 0.6800 & 1.0000 & N.S. \\
            & 2 & 100 & 6.00  & 10.33 & 4.33 & [-0.67, 10.00] & 0.1353 & 1.0000 & N.S. \\
            & 3 & 100 & 4.00  & 3.67  & -0.33 & [-3.00, 2.67] & 1.0000 & 1.0000 & N.S. \\
            & 4 & 100 & 3.67  & 4.33  & 0.67 & [-2.67, 4.33] & 0.8579 & 1.0000 & N.S. \\
\bottomrule
\end{tabular}

\vspace{5pt}

\begin{minipage}{\linewidth}
\small
\textbf{Note:} Five $\text{model} \times \text{level} \times \text{method}$ cells include fewer than 300 attempts because of the exclusions described in \textit{Evaluation Protocol and Metrics} section: GPT-5-mini FISER at Levels 1 and 3, GPT-5-mini CEMIR at Levels 3 and 4, and GPT-5-nano CEMIR at Level 4.
\end{minipage}

\end{table}

\FloatBarrier
\begin{figure}[!ht]
\centering
\includegraphics[width=\textwidth]{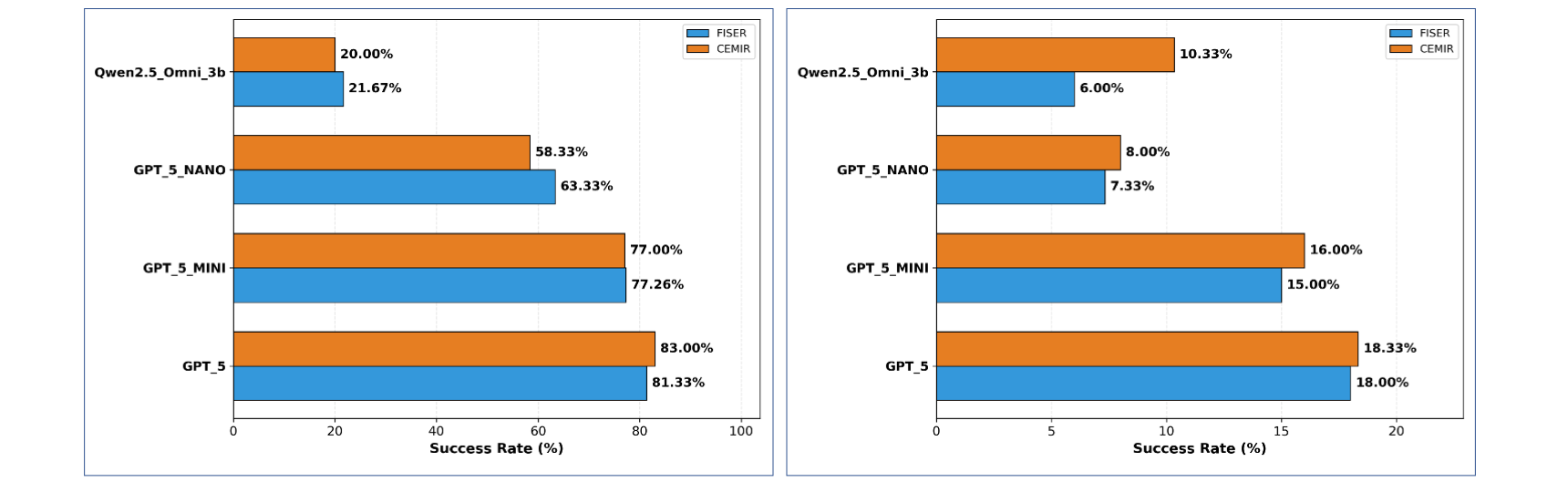}
\caption{Model-specific comparison of Success Rates between FISER and CEMIR for Level 1 (left) and Level 2 (right).}
\label{Model-specific comparison1}
\end{figure}

\begin{figure}[!ht]
\centering
\includegraphics[width=\textwidth]{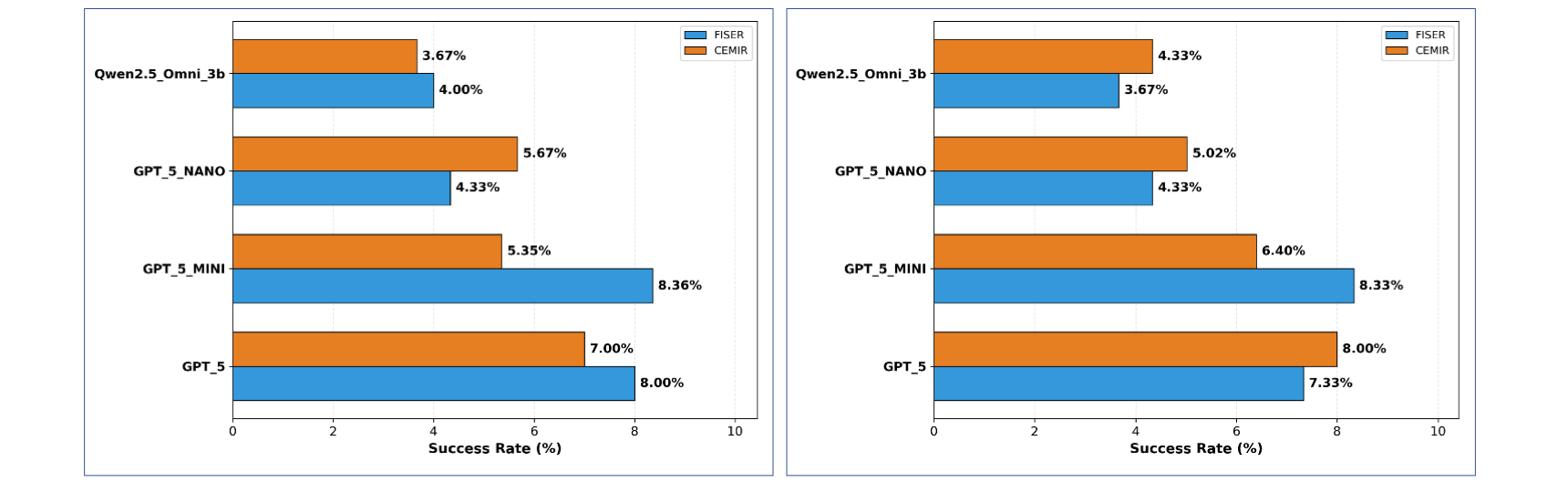}
\caption{Model-specific comparison of Success Rates between FISER and CEMIR for Level 3 (left) and 
Level 4 (right).}
\label{Model-specific comparison2}
\end{figure}

\begin{table} [!ht]
\centering
\small
\setlength{\tabcolsep}{3pt}
\caption{Ablation study on Qwen2.5‑Omni‑3B: Success Rate (SR\%) of FISER vs. two modality conditions (Raw Text + Generated Narrative; Raw Text + Image) across four difficulty levels. $\Delta$ = Modality SR $-$ FISER SR. N = 100 episodes per cell.}
\label{table4}
\begin{tabular}{
  >{\centering\arraybackslash}p{0.8cm}
  >{\raggedright\arraybackslash}p{2.3cm}
  >{\centering\arraybackslash}p{0.7cm}
  >{\centering\arraybackslash}p{1.2cm}
  >{\centering\arraybackslash}p{1.2cm}
  >{\centering\arraybackslash}p{0.9cm}
  >{\centering\arraybackslash}p{1.6cm}
  >{\centering\arraybackslash}p{1.1cm}
  >{\centering\arraybackslash}p{1.1cm}
  >{\raggedright\arraybackslash}p{1.1cm}
}
\toprule
\textbf{Level} & \textbf{Modality} & \textbf{n} & \textbf{FISER SR (\%)} & \textbf{Modality SR (\%)} & \textbf{$\Delta$ (pp)} & \textbf{95\% CI} & \textbf{$p$-value} & \textbf{$p$-value (Holm)} & \textbf{Sig.} \\
\midrule
1 & Raw Text + Generated Narrative & 100 & 21.67 & 27.00 &  5.33 & [-0.67, 11.33] & 0.0967 & 0.5799 & N.S. (Not Significant) \\
  & Raw Text + Image               & 100 & 21.67 & 16.67 & -5.00 & [-12.00, 2.00] & 0.1861 & 0.6078 & N.S. \\
\cmidrule(lr){1-10}
2 & Raw Text + Generated Narrative & 100 &  6.00 &  9.67 &  3.67 & [-0.67, 8.33]  & 0.1408 & 0.6078 & N.S. \\
  & Raw Text + Image               & 100 &  6.00 & 13.33 &  7.33 & [2.00, 12.67]  & 0.0090 & 0.0716 & N.S. \\
\cmidrule(lr){1-10}
3 & Raw Text + Generated Narrative & 100 &  4.00 &  6.00 &  2.00 & [-1.00, 5.33]  & 0.3067 & 0.6078 & N.S. \\
  & Raw Text + Image               & 100 &  4.00 &  2.00 & -2.00 & [-4.33, 0.34]  & 0.2140 & 0.6078 & N.S. \\
\cmidrule(lr){1-10}
4 & Raw Text + Generated Narrative & 100 &  3.67 &  7.33 &  3.67 & [0.00, 7.67]   & 0.1216 & 0.6078 & N.S. \\
  & Raw Text + Image               & 100 &  3.67 &  8.00 &  4.33 & [0.00, 8.67]   & 0.0641 & 0.4487 & N.S. \\
\bottomrule
\end{tabular}

\vspace{5pt}

\begin{minipage}{\linewidth}
\small
\textbf{Note:} The \texttt{p\_value} column reports unadjusted $p$-values from the cluster-level paired permutation test; Holm $p$ reports Holm--Bonferroni-adjusted values across the 8 tests in this table. The 95\% CIs are cluster-bootstrap intervals and are not adjusted for multiple comparisons; a CI excluding zero (e.g., Level~2, Raw Text + Image) can therefore accompany a non-significant Holm-adjusted $p$ ($> 0.05$). Significance (Sig.) is based on the Holm-adjusted $p$.
\end{minipage}

\end{table}

This is consistent with HandMeThat's textual interface, which leaves no physical information withheld from a text-only model \cite{wan2022handmethat}: the information needed to resolve ambiguity is, in principle, recoverable from text alone. Recovering it in practice, however, still requires the right kind of processing over that text. FISER's decomposition into explicit social and embodied reasoning stages already improves substantially over weaker, less structured text-only baselines on this benchmark \cite{Wan2025}, indicating that structured reasoning, not the mere presence of information in the text, is what closes most of the gap. Our finding of no statistically significant difference between FISER and CEMIR extends this observation one step further: augmenting the episode text with generated narrative and visual content, in place of FISER's explicit reasoning stages, conferred no measurable advantage over structured reasoning applied to the original text. For instance, when the instruction is "hand me the dusty one," the agent must identify the correct referent among several candidates by linking the linguistic descriptor to the appropriate object given the scene context and preceding actions; this requires reasoning over the text, not additional content, and Figure 6 summarizes this bottleneck conceptually. \newline 

Regarding Winston's Strong Story and Directed Perception Hypotheses \cite{winston2011strong}, which motivated CEMIR's combination of narrative and visual content as an operationalization of reasoning over "imagined events": this result does not speak to the validity of those hypotheses as accounts of human cognition, which lie outside the scope of this evaluation. It does indicate that, in this specific implementation and benchmark, generating such content from the episode's own text yielded no measurable value beyond what FISER's structured reasoning already extracted from that same text. It is also worth noting that CEMIR's enrichment pipeline requires an additional language model (Qwen3-8B) and an additional diffusion model (FLUX) to generate a narrative and image for every action step, whereas FISER operates directly on the textual episode with no additional model calls; this added computational cost did not correspond to any measurable gain in accuracy in our evaluation. \newline

\begin{figure}[H]
\centering
\includegraphics[width=\textwidth]{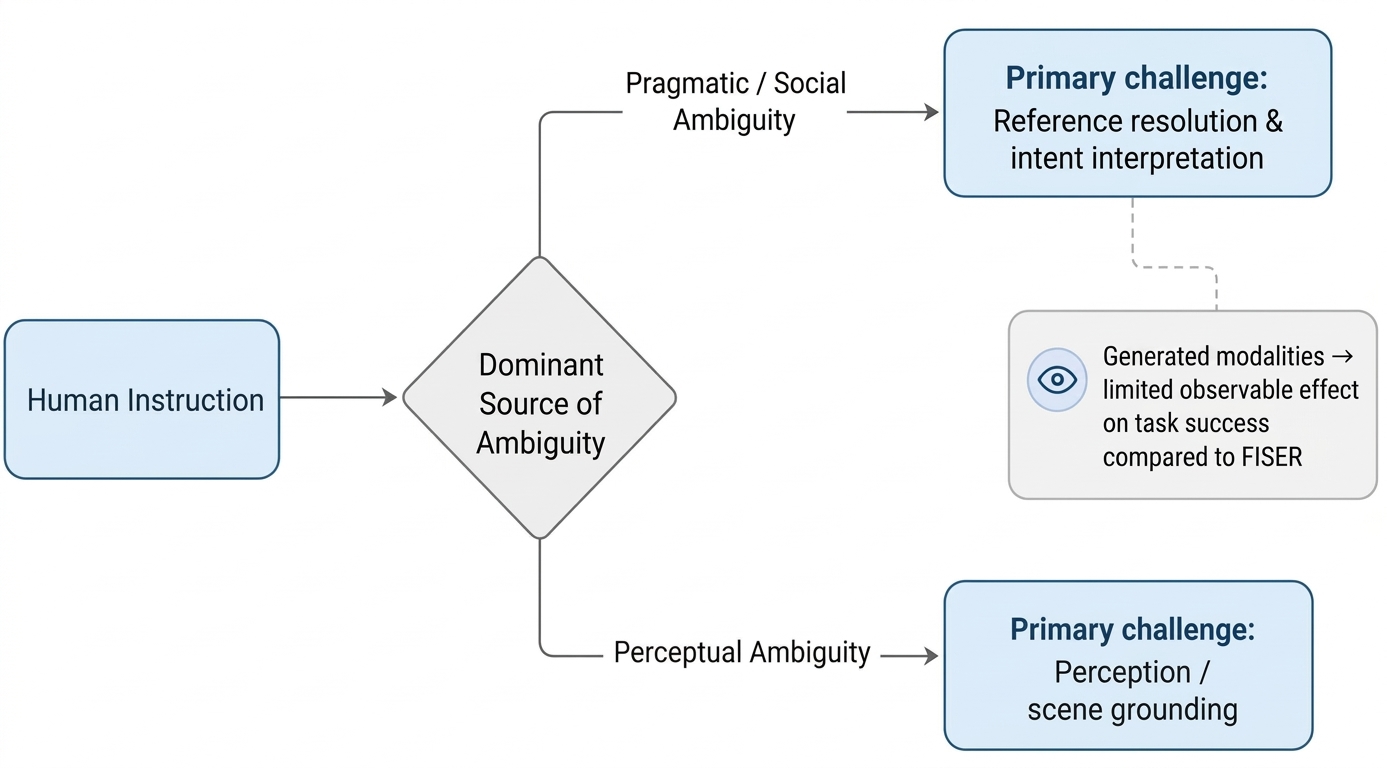}
\caption{Conceptual framework illustrating the primary task bottlenecks.}
\label{Conceptual framework}
\end{figure}

The finding above should be scoped carefully. Our comparison isolates enrichment that is synthesized entirely from the episode text, with no independent access to the environment; it does not speak to enrichment grounded in directly observed perceptual signals, as in MMC-GAN's use of the robot's physical state and scene clutter \cite{magassouba2018multimodal}, SemanticScanpath's use of gaze behavior \cite{ menendez2025semanticscanpath}, or PaLM-E's integration of real sensor observations \cite{ driess2023palme}. In HandMeThat v2, this latter form of grounding is structurally unavailable: the benchmark's textual interface already states every object and its relevant properties, so there is no independently observable signal left for a perceptual channel to supply \cite{wan2022handmethat}. Our null result therefore constrains only the self-generated branch of enrichment-based approaches; whether directly observed perceptual grounding would fare differently under pragmatic ambiguity, in a setting where such grounding is actually available, remains untested and is a natural extension for future work using embodied benchmarks that retain genuine visual observability. \newline

Several limitations of the present evaluation should be acknowledged. First, as discussed above, the experiments are conducted exclusively on HandMeThat v2, whose ambiguity is predominantly pragmatic rather than perceptual; the conclusions drawn here are correspondingly scoped and may not extend to benchmarks where perceptual uncertainty is the dominant source of difficulty. Second, the proposed enrichment strategy is evaluated under a single architectural configuration. Alternative multimodal integration strategies, different perception modules, or richer visual encodings may lead to different outcomes, and exploring these variations may clarify when and how multimodal representations contribute to embodied task performance. Third, while a targeted ablation study was conducted on Qwen2.5‑Omni‑3B to examine the separate contributions of the visual and narrative modalities (Ablation Study section), computational constraints precluded extending this analysis to the larger backbones evaluated elsewhere in this work; whether the same null pattern holds at larger model scales remains untested. Fourth, while a manual fidelity check (Content Fidelity Verification section) found that generated content was consistent with ground-truth episode state in 94\% of audited action steps, this check was necessarily limited to a stratified sample rather than the full dataset, and residual inaccuracies in the unaudited majority, or in the small fraction of failed cases, cannot be entirely ruled out. We consider it unlikely that such inaccuracies account for the observed null result, given the high fidelity rate and consistency of this result across four difficulty levels, but a fully automated or exhaustive fidelity audit was beyond the scope of this study. Fifth, our comparison is deliberately conducted at the paradigm level: FISER and CEMIR are each evaluated as a complete system, with the prompting strategy that characterizes its approach, mirroring how the two paradigms would be deployed in practice. The results therefore address which approach is preferable under pragmatic ambiguity, rather than the isolated contribution of any single component. Attributing performance separately to enrichment and to explicit reasoning would require a factorial design, which we identify as a natural extension of this work.\newline

Future work may address these limitations directly. Evaluating multimodal representations on benchmarks where perceptual ambiguity plays a more prominent role, for example, tasks involving visually similar objects, partial observability, or incomplete scene information, would provide a more direct test of whether enriched perceptual representations improve task success under the conditions they are typically motivated by. Extending the modality ablation to larger backbones, and to benchmarks with higher baseline success rates, would help disentangle a genuine level-specific effect from an artifact of low-baseline statistics. Exploring alternative multimodal fusion strategies or richer perceptual encodings may further reveal conditions under which multimodal integration becomes advantageous, and more detailed analysis of agent decision processes and failure cases may help clarify how different sources of ambiguity affect embodied task performance. Overall, this study's central finding is a negative but informative one: for pragmatic, intent-based ambiguity in a text-complete environment, an enrichment-based system built on generated narrative and imagery did not outperform a reasoning-based system built on structured inference over the same text, at any evaluated model scale or difficulty level. This closes the gap identified in Research Gap section: under this specific form of ambiguity, and within the scope of HandMeThat v2 and CEMIR's implementation, an enrichment-based alternative did not achieve a measurable advantage over structured text-only reasoning. Whether this reflects a general limitation of self-generated enrichment under pragmatic ambiguity, or a property specific to CEMIR's design or to HandMeThat v2, remains for the future work outlined above to determine.

\section*{Conclusion}

This study compared two paradigms for resolving pragmatic, intent-based ambiguity on HandMeThat v2: FISER, a reasoning-based approach that explicitly decomposes intent recognition into social and embodied reasoning stages over episode text, and CEMIR, an enrichment-based approach that augments that same text with generated narrative and imagery. Evaluating both across four language model backbones and four difficulty levels, we found no statistically significant difference between the two approaches in any of the sixteen model-by-difficulty comparisons: CEMIR's performance was statistically indistinguishable from FISER's throughout, with success rate driven primarily by backbone capability and task difficulty rather than by the choice of paradigm. Within the scope of this benchmark and this implementation, an enrichment-based alternative built on self-generated narrative and visual content did not achieve a measurable advantage over structured text-only reasoning. A targeted ablation on Qwen2.5‑Omni‑3B examined the narrative and visual components individually. Across this and the main comparison, 32 statistical tests in total, with Holm-Bonferroni correction applied independently within each table (families of 4, 4, 16, and 8 tests for Tables 1–4, respectively), no result reached significance. These findings are scoped specifically to enrichment synthesized entirely from an episode's own text, evaluated against a strong structured-reasoning baseline, on a benchmark where all task-relevant information is available in that text. They do not speak to enrichment grounded in directly observed perceptual signals, nor to settings where ambiguity arises from genuine perceptual uncertainty rather than from inferring a person's unstated intent. Future work extending this comparison to larger backbones, alternative fusion strategies, and benchmarks with perceptually grounded ambiguity would help establish how far this result generalizes beyond the specific paradigm, benchmark, and implementation examined here.

\section*{Data availability}

The code used for evaluation and statistical analysis is available at:
\url{https://github.com/CEMIR-Project/CEMIR}. The enrichment dataset, including the generated narratives and images, is available at:
\url{https://huggingface.co/datasets/CEMIR-Project/HandMeThat-V2-CEMIR}. The HandMeThat v2 benchmark used as the base dataset is publicly available at: \url{https://github.com/Simon-Wan/HandMeThat}.

\section*{Funding}

The authors received no specific funding for this work.

\section*{Competing interests}

The authors have declared that no competing interests exist.

\section*{Author contributions (CRediT)}

\textbf{Ziya Hosseynzadeh:} Conceptualization, Data curation, Formal analysis, Investigation, Methodology, Project administration, Validation, Visualization, Writing -- original draft, Writing -- review \& editing. \textbf{Farjad Fallah:} Data curation, Formal analysis, Investigation, Methodology, Software, Validation, Visualization. \textbf{Farnoosh Fallah:} Data curation, Formal analysis, Investigation, Methodology, Software, Validation, Visualization. \textbf{Soroush Gooran:} Conceptualization, Formal analysis, Project administration, Supervision, Validation, Writing -- review \& editing. \textbf{Hossein Sameti:} Conceptualization, Formal analysis, Investigation, Project administration, Supervision. \textbf{Mohammad Amanloo:} Data curation, Formal analysis, Software.

\section*{Ethics statement}

This study did not involve human participants, animals, or personally identifiable data.

\section*{Use of generative AI}

Generative AI tools Claude (Anthropic) and ChatGPT (OpenAI) were used solely to improve the readability and correct the grammar of the manuscript text. Figures 1 and 6 were conceived and designed by the authors, who specified their complete content and structure; the Gemini (Google) image generation tool was used only to render the final graphics. The authors reviewed and verified all text and all generated figures and take full responsibility for the content of the manuscript.

\bibliographystyle{unsrt}
\bibliography{references}

@article{vanhorenbeke2021activity,
  author  = {Van-Horenbeke, Franz A. and Peer, Angelika},
  title   = {Activity, Plan, and Goal Recognition: A Review},
  journal = {Frontiers in Robotics and AI},
  volume  = {8},
  pages   = {643010},
  year    = {2021},
  doi     = {10.3389/frobt.2021.643010},
  url     = {https://www.frontiersin.org/articles/10.3389/frobt.2021.643010}
}

@article{zhou2024enhancing,
  author  = {Zhou, Jincao and Su, Xuezhong and Fu, Weiping and Lv, Yang and Liu, Bo},
  title   = {Enhancing intention prediction and interpretability in service robots with LLM and KG},
  journal = {Scientific Reports},
  year    = {2024},
  volume  = {14},
  number  = {26999},
  doi     = {10.1038/s41598-024-77916-3},
  url     = {https://www.nature.com/articles/s41598-024-77916-3}
}

@article{chandrakala2024intent,
author = {Chandrakala, C. B. and Bhardwaj, Rohit and Pujari, Chetana},
title = {An intent recognition pipeline for conversational AI},
journal = {International Journal of Information Technology},
year = {2024},
volume = {16},
number = {2},
pages = {731--743},
doi = {10.1007/s41598-023-01642-8}
}

@article{benayas2024enhancing,
author = {Benayas, Alberto and Miguel-Ángel, Sicilia and Mora-Cantallops, Marçal},
title = {Enhancing Intent Classifier Training with Large Language Model-generated Data},
journal = {Applied Artificial Intelligence},
year = {2024},
volume = {38},
number = {1},
pages = {2414483},
doi = {10.1080/08839514.2024.2414483}
}

@book{suchman1987plans,
  author    = {Lucy A. Suchman},
  title     = {Plans and Situated Actions: The Problem of Human-Machine Communication},
  year      = {1987},
  publisher = {Cambridge University Press},
  address   = {Cambridge}
}

@incollection{clark1991grounding,
  author    = {Clark, Herbert H. and Brennan, Susan E.},
  title     = {Grounding in Communication},
  booktitle = {Perspectives on Socially Shared Cognition},
  editor    = {Resnick, Lauren B. and Levine, John M. and Teasley, Stephanie D.},
  publisher = {American Psychological Association},
  address   = {Washington, DC},
  year      = {1991},
  pages     = {127--149}
}

@article{lemaignan2012grounding,
  author={Lemaignan, S{\'e}verin and Ros, Raquel and Sisbot, E. Akin and Alami, Rachid and Beetz, Michael},
  journal={International Journal of Social Robotics}, 
  title={Grounding the Interaction: Anchoring Situated Discourse in Everyday Human-Robot Interaction}, 
  year={2012},
  volume={4},
  number={2},
  pages={181--199},
  doi={10.1007/s12369-011-0123-x}
}

@inproceedings{Wan2025,
  author       = {Yanming Wan and Yue Wu and Yiping Wang and Jiayuan Mao and Natasha Jaques},
  title        = {Infer Human’s Intentions Before Following Natural Language Instructions},
  booktitle    = {Proceedings of the Thirty-Ninth AAAI Conference on Artificial Intelligence (AAAI-25)},
  year         = {2025},
  organization = {Association for the Advancement of Artificial Intelligence},
  url          = {https://arxiv.org/abs/2409.18073},
  note         = {Extended version available at \url{https://arxiv.org/abs/2409.18073}},
  code         = {https://github.com/Simon-Wan/FISER}
}

@inproceedings{wan2022handmethat,
  author    = {Wan, Yanming and Mao, Jiayuan and Tenenbaum, Joshua B.},
  title     = {HandMeThat: Human-Robot Communication in Physical and Social Environments},
  booktitle = {Advances in Neural Information Processing Systems (NeurIPS) Track on Datasets and Benchmarks},
  volume    = {35},
  pages     = {1--13},
  year      = {2022}
}

@inproceedings{tan2020vokenization,
  title     = {Vokenization: Improving Language Understanding with Contextualized, Visual-Grounded Supervision},
  author    = {Tan, Hao and Bansal, Mohit},
  booktitle = {Proceedings of the 2020 Conference on Empirical Methods in Natural Language Processing (EMNLP)},
  pages     = {2066--2080},
  year      = {2020}
}

@inproceedings{yang2022zlavi,
  title     = {{Z-LaVI}: Zero-Shot Language Solver Fueled by Visual Imagination},
  author    = {Yang, Yue and Yao, Wenlin and Zhang, Hongming and Wang, Xiaoyang and Yu, Dong and Chen, Jianshu},
  booktitle = {Proceedings of the 2022 Conference on Empirical Methods in Natural Language Processing (EMNLP)},
  pages     = {1186--1203},
  year      = {2022},
  publisher = {Association for Computational Linguistics},
  doi       = {10.18653/v1/2022.emnlp-main.78},
  url       = {https://aclanthology.org/2022.emnlp-main.78}
}

@inproceedings{lu2022imagination,
    title = "Imagination-Augmented Natural Language Understanding",
    author = "Lu, Yujie  and
      Zhu, Wanrong  and
      Wang, Xin Eric  and
      Eckstein, Miguel  and
      Wang, William Yang",
    booktitle = "Proceedings of the 2022 Conference of the North American Chapter of the Association for Computational Linguistics: Human Language Technologies",
    month = jul,
    year = "2022",
    address = "Seattle, United States",
    publisher = "Association for Computational Linguistics",
    url = "https://aclanthology.org/2022.naacl-main.325",
    doi = "10.18653/v1/2022.naacl-main.325",
    pages = "4392--4402"
}

@article{huang2026synthetic,
  title={Synthetic Perception: Can Generated Images Unlock Latent Visual Prior for Text-Centric Reasoning?},
  author={Huang, Yuesheng and Zhang, Peng and Wu, Xiaoxin and Liu, Riliang and Liang, Jiaqi},
  journal={arXiv preprint arXiv:2506.17623v2},
  year={2026}
}

@inproceedings{winston2011strong,
  author    = {Winston, Patrick Henry},
  title     = {The Strong Story Hypothesis and the Directed Perception Hypothesis},
  booktitle = {AAAI Fall Symposium Series},
  year      = {2011},
  pages     = {n. pag.},
  publisher = {Association for the Advancement of Artificial Intelligence},
  url       = {http://hdl.handle.net/1721.1/67693}
}

@article{Byom2013ToM,
  author  = {Lindsey J. Byom and Bilge Mutlu},
  title   = {Theory of mind: mechanisms, methods, and new directions},
  journal = {Frontiers in Human Neuroscience},
  year    = {2013},
  volume  = {7},
  pages   = {413},
  month   = aug,
  doi     = {10.3389/fnhum.2013.00413}
}

@article{Baker2017ToM,
  author  = {Chris L. Baker and Julian Jara-Ettinger and Rebecca Saxe and Joshua B. Tenenbaum},
  title   = {Rational quantitative attribution of beliefs, desires and percepts in human mentalizing},
  journal = {Nature Human Behaviour},
  volume  = {1},
  pages   = {0064},
  year    = {2017},
  doi     = {10.1038/s41562-017-0064}
}

@article{Baker2011BToM,
  author    = {Baker, Chris L. and Saxe, Rebecca and Tenenbaum, Joshua B.},
  title     = {Bayesian Theory of Mind: Modeling Joint Belief-Desire Attribution},
  journal   = {Cognitive Science},
  volume    = {35},
  number    = {2},
  pages     = {246--279},
  year      = {2011},
}

@inproceedings{zhixuan2022solving,
  author    = {Zhi-Xuan, Tan and Gothoskar, Nishad and Pollok, Falk and Gutfreund, Dan and Tenenbaum, Joshua B. and Mansinghka, Vikash K.},
  title     = {Solving the Baby Intuitions Benchmark with a Hierarchically Bayesian Theory of Mind},
  booktitle = {Proceedings of the IEEE Conference on Artificial Intelligence (CAI)},
  year      = {2022},
  pages     = {1--6},
  publisher = {IEEE},
  note      = {arXiv preprint arXiv:2208.02914}
}

@inproceedings{Bandyopadhyay2012,
  author    = {Bandyopadhyay, Tirthankar and Won, Kok Sung and Frazzoli, Emilio and Hsu, David and Lee, Wee Sun and Rus, Daniela},
  title     = {Intention-Aware Motion Planning},
  booktitle = {Proceedings of the International Workshop on the Algorithmic Foundations of Robotics (WAFR)},
  year      = {2012},
  pages     = {1--16}
}

@inproceedings{tan2023language,
  author    = {Tan, Zhi-Xuan and Lunis, Paul Stefan and Fernandez Echeverri, Nathalie and Mansinghka, Vikash and Tenenbaum, Josh},
  title     = {Language Models as Informative Goal Priors in a Bayesian Theory of Mind},
  booktitle = {Proceedings of the Annual Meeting of the Cognitive Science Society},
  volume    = {45},
  number    = {45},
  pages     = {3854--3860},
  year      = {2023},
  publisher = {Cognitive Science Society}
}

@inproceedings{zhixuan2024clips,
  author    = {Zhi-Xuan, Tan and Ying, Lance and Mansinghka, Vikash and Tenenbaum, Joshua B.},
  title     = {Pragmatic Instruction Following and Goal Assistance via Cooperative Language-Guided Inverse Planning},
  booktitle = {Proceedings of the 23rd International Conference on Autonomous Agents and Multiagent Systems (AAMAS 2024)},
  pages     = {2094--2103},
  year      = {2024},
  publisher = {IFAAMAS}
}

@article{saad2025theory,
  title     = {Theory of Mind in Action: The Instruction Inference Task in Dynamic Human-Agent Collaboration},
  author    = {Saad, Fardin and Murukannaiah, Pradeep K. and Singh, Munindar P.},
  journal   = {arXiv preprint arXiv:2507.02935},
  year      = {2025}
}

@article{hwang2025maskedirl,
  title={Masked IRL: LLM-Guided Reward Disambiguation from Demonstrations and Language},
  author={Hwang, Minyoung and Forsey-Smerek, Alexandra and Dennler, Nathaniel and Bobu, Andreea},
  journal={arXiv preprint arXiv:2511.14565},
  year={2025}
}

@inproceedings{tellex2011understanding,
  author    = {Tellex, Stefanie and Kollar, Thomas and Dickerson, Steven and Walter, Matthew R. and Banerjee, Ashis Gopal and Teller, Seth and Roy, Nicholas},
  title     = {Understanding Natural Language Commands for Robotic Navigation and Mobile Manipulation},
  booktitle = {Proceedings of the National Conference on Artificial Intelligence (AAAI 2011)},
  year      = {2011},
  pages     = {1507--1514},
  address   = {San Francisco, CA, USA}
}

@article{magassouba2018multimodal,
  author={Magassouba, Aly and Sugiura, Komei and Kawai, Hisashi},
  journal={IEEE Robotics and Automation Letters}, 
  title={A Multimodal Classifier Generative Adversarial Network for Carry and Place Tasks From Ambiguous Language Instructions}, 
  year={2018},
  volume={3},
  number={4},
  pages={3115-3122},
  doi={10.1109/LRA.2018.2849607}
}

@article{menendez2025semanticscanpath,
  title     = {SemanticScanpath: Combining Gaze and Speech for Situated Human-Robot Interaction Using LLMs},
  author    = {Menendez, Elisabeth and Gienger, Michael and Mart{\'\i}nez, Santiago and Balaguer, Carlos and Belardinelli, Anna},
  journal   = {arXiv preprint arXiv:2503.16548},
  year      = {2025}
}

@inproceedings{driess2023palme,
  author    = {Driess, Danny and Xia, Fei and Sajjadi, Mehdi S. M. and Lynch, Corey and Chowdhery, Aakanksha and Ichter, Brian and Wahid, Ayzaan and Tompson, Jonathan and Vuong, Quan and Yu, Tianhe and Huang, Wenlong and Chebotar, Yevgen and Sermanet, Pierre and Duckworth, Daniel and Levine, Sergey and Vanhoucke, Vincent and Hausman, Karol and Toussaint, Marc and Greff, Klaus and Zeng, Andy and Mordatch, Igor and Florence, Pete},
  title     = {{PaLM-E}: An Embodied Multimodal Language Model},
  booktitle = {Proceedings of the 40th International Conference on Machine Learning (ICML)},
  series    = {Proceedings of Machine Learning Research},
  volume    = {202},
  pages     = {1--20},
  year      = {2023},
  publisher = {PMLR}
}

@article{harnad1990symbol,
  title={The symbol grounding problem},
  author={Harnad, Stevan},
  journal={Physica D: Nonlinear Phenomena},
  volume={42},
  number={1-3},
  pages={335--346},
  year={1990},
  publisher={Elsevier}
}

@article{qwen3, title = {Qwen3 Technical Report}, author = {{Qwen Team}}, journal = {arXiv preprint arXiv:2505.09388}, year = {2025}, url = {https://arxiv.org/abs/2505.09388} }

@article{qwen25omni, title = {Qwen2.5-Omni Technical Report}, author = {Xu, Jin and Guo, Zhifang and He, Jinzheng and Hu, Hangrui and He, Ting and Bai, Shuai and Chen, Keqin and Wang, Jialin and Fan, Yang and Dang, Kai and Zhang, Bin and Wang, Xiong and Chu, Yunfei and Lin, Junyang}, journal = {arXiv preprint arXiv:2503.20215}, year = {2025}, url = {https://arxiv.org/abs/2503.20215} }

@misc{flux2024,
  author       = {{Black Forest Labs}},
  title        = {{FLUX.1 [dev]}},
  year         = {2024},
  howpublished = {Hugging Face model repository},
  url          = {https://huggingface.co/black-forest-labs/FLUX.1-dev}
}

@inproceedings{bisk2020experience,
  title     = {Experience Grounds Language},
  author    = {Bisk, Yonatan and Holtzman, Ari and Thomason, Jesse and Andreas, Jacob and Bengio, Yoshua and Chai, Joyce and Lapata, Mirella and Lazaridou, Angeliki and May, Jonathan and Nisnevich, Aleksandr and Pinto, Nicolas and Turian, Joseph},
  booktitle = {Proceedings of the 2020 Conference on Empirical Methods in Natural Language Processing (EMNLP)},
  pages     = {8718--8735},
  year      = {2020},
  address   = {Online},
  publisher = {Association for Computational Linguistics},
  doi       = {10.18653/v1/2020.emnlp-main.703}
}
\end{document}